\documentclass{new_tlp}

\usepackage{url}
\usepackage{dsfont,amsmath}
\usepackage{mathtools}
\usepackage{amsfonts}
\usepackage{amssymb}
\usepackage{xspace}
\usepackage{color}
\usepackage{comment}
\usepackage[ruled,linesnumbered,vlined]{algorithm2e}
\usepackage{todonotes}
\usepackage{enumerate}
\usepackage{amssymb}

\newtheorem{Example}{Example}

\title[On the Configuration of More and Less Expressive Logic Programs]{On the Configuration of More and Less Expressive Logic Programs\thanks{Mauro Vallati was supported by a UKRI Future Leaders Fellowship [grant number MR/T041196/1].}}

\author[Dodaro, Maratea, and Vallati] {
CARMINE DODARO\\
University of Calabria, Italy\\
\email{dodaro@mat.unical.it}
\and
MARCO MARATEA\\
University of Genoa, Italy\\
\email{marco.maratea@unige.it}
\and 
MAURO VALLATI\\
University of Huddersfield, UK\\
\email{m.vallati@hud.ac.uk}
}

\jdate{July 2020}
\pubyear{2020}
\pagerange{\pageref{firstpage}--\pageref{lastpage}}
\doi{S1471068401001193}

\begin{document}

\label{firstpage}
\maketitle

\begin{abstract}
The decoupling between the representation of a certain problem, i.e., its knowledge model, and the reasoning side is one of main strong points of model-based Artificial Intelligence (AI). This allows, e.g. to focus on improving the reasoning side by having advantages on the whole solving process. Further, it is also well-known that many  solvers are very sensitive to even syntactic changes in the input.

In this paper, we focus on improving the reasoning side by taking advantages of such sensitivity. We consider two well-known model-based AI methodologies, SAT and ASP, define a number of syntactic features that may characterise their inputs, and use automated configuration tools to reformulate the input formula or program. Results of a wide experimental analysis involving SAT and ASP domains, taken from respective competitions, show the different advantages that can be obtained by using input reformulation and configuration. 
Under consideration in Theory and Practice of Logic Programming (TPLP).
\end{abstract}


\section{Introduction}
Model-based reasoning is one of the most prominent areas of research in Artificial Intelligence (AI). 
In model-based AI approaches, solvers accept input instances written in a given logical language and  automatically  compute  their  solutions \cite{DBLP:conf/ijcai/Geffner18}.

A pillar of model-based AI is the decoupling between the knowledge model and the reasoning side, which is usually referred to as domain-independent reasoning. This supports the use of knowledge engineering approaches that separate the modelling part from the reasoning part. The main advantage of  such separation is that it is possible to ``optimise" one of the two parts without changing the other for obtaining overall advantage in the whole process. 

We follow this path, further evidencing that this modular approach also supports the use of reformulation and configuration techniques which can automatically re-formulate, re-represent or tune the knowledge model, while keeping the same input language, in order to increase the efficiency of a general solver (see, e.g. \cite{DBLP:conf/ijcai/VallatiHCM15} still for the case of automated planning, where reformulation techniques have been widely applied). The idea is to make these techniques to some degree independent of domain knowledge and solver (that is, applicable to a range of domains and solvers technology), and use them to form a wrapper around a solver, improving its overall performance for the problem to which it is applied.

In this paper, we investigate how the configuration of knowledge models, i.e., the order in which elements are listed in the considered model, can affect the performance of general automated solvers in the wider field of logic programming. In particular, we focus on two areas that can be considered to end of the spectrum: Propositional Satisfiability (SAT) and Answer Set Programming (ASP) \cite{baral2003,DBLP:conf/iclp/GelfondL88,DBLP:journals/ngc/GelfondL91,DBLP:journals/cacm/BrewkaET11}. In both areas, configuration has been traditionally exploited to modify the behaviour of solvers to improve performance on a considered (class of) instance(s)  \cite{DBLP:journals/jair/EggenspergerLH19}. With regard to the knowledge models characteristics, SAT, particularly in its CNF connotation, has a limited expressivity in terms of the syntax of the models. On the other hand, ASP has instead a great level of expressivity, having a rich syntax \cite{DBLP:journals/tplp/CalimeriFGIKKLM20} aiming for better readability.

In fact, the two approaches, while sharing similar aspects, e.g. (i) presence of a somehow similar input format for the propositional part, (ii) the reasoning part of state-of-the-art SAT and ASP solvers employ variations of the CDCL algorithm \cite{DBLP:journals/eatcs/Mitchell05,DBLP:journals/ai/GebserKS12}, and (iii) the existence of linear techniques for rewriting ASP programs falling in a certain syntactic class (cf. tight, \cite{DBLP:journals/tplp/ErdemL03}) and solvers that exploit such property to use SAT as ASP solver, are also very different on several respects. Among others: (a) ASP is a first order language allowing for variables, that are eliminated during grounding, and during grounding some kind of reformulation already happens, (b) ASP rules are somehow more ``constrained" than CNF clauses, since they need to preserve the head-body structure, (c) ASP allows for a number of additional constructs, like aggregates, and (d) propositional ASP is strictly more expressive than SAT. 



Building on the experience gained in our prior work on SAT \cite{DBLP:conf/aiia/VallatiM19}, the large experimental analysis presented in this paper provides a collection of results that help to understand the impact of knowledge model configuration on automated solvers from these two subareas of the logic programming field, and to provide valuable support to knowledge engineers. In particular, in this work we:
\begin{enumerate}
    \item define a number of SAT and ASP syntactic features useful for analysing the structure of the formula/program at hand,
    \item introduce a framework that, leveraging on the introduced features, allows the automated reconfiguration of the input formula/program,  
    \item employ SMAC~\cite{DBLP:conf/lion/HutterHL11} as a configuration tool to reformulate the input formula/program, and 
    \item compare the performance of state-of-the-art SAT and ASP solvers on basic and reformulated formulae/programs coming from well-known benchmark domains taken for respective competitions (see. e.g. the reports of the last competitions available \cite{DBLP:journals/jsat/HeuleJS19,DBLP:journals/tplp/GebserMR20}.
\end{enumerate}

Results show that SAT solvers can greatly benefit from such reformulation, being able to solve a consistent number of additional instances or in shorter time, and that same happens to ASP solvers, to a lesser degrees, despite the limitations (a) and (b), and the wider degrees of parameters to analyse, cf. (c).



This paper is organised as follows. First, in Section \ref{sec:prel}, we present needed preliminaries about SAT and ASP, their input languages, and the configuration techniques we are going to exploit in the paper. Then, Section \ref{sec:know} is devoted to the configuration of SAT formulae and ASP programs, by defining the syntactic features we are going to employ for reformulation. Further, Section \ref{sec:exp} presents the results of our experimental analysis on both SAT and ASP domains. The paper ends in Section \ref{sec:rel} and \ref{sec:conc} with the analysis of related literature and conclusions, respectively. 

\section{Preliminaries}
\label{sec:prel}

This section provides the essential background with regard to SAT and ASP fields, and with regard to automated configuration techniques.

\subsection{SAT Formulae and Answer Set Programming}

We define (ground) disjunctive ASP programs and SAT formulae so as to underline similarities, in order to make it easier in later sections to compare the presented techniques.

\paragraph{Syntax.}
Let $\mathcal{A}$ be a propositional signature.
An element $p \in \mathcal{A}$ is called \textit{atom} or \emph{positive literal}. The negation of an atom $p$, in symbols $\lnot p$, is called \emph{negative literal}.
Given a literal $l$, we define $\overline{l}=p$, if $l=p$ and $\overline{l}=\neg p$, if $l=p$ for some $p\in\mathcal{A}$.
Given a set of literals $M$, we denote by $M^+$ the set of positive literals of
$M$, by $M^-$ the set of negative literals of $M$, and by $\overline{M}$ the set $\{\overline{l}\ :\  l\in M\}$.
A \emph{clause} is a finite set of literals (seen as a disjunction), and a \emph{SAT formula} is a finite set of clauses (seen as a conjunction).

\begin{Example}\label{ex:cnf}
    Let $\varphi_{run}$ be the following SAT formula:
    \begin{align*}
    \begin{array}{ll}
    c_1:& \{p_1, \lnot p_3\} \\
    c_2:& \{p_2, p_3, \lnot p_1, \lnot p_4\} \\
    c_3:& \{\lnot p_5, \lnot p_4\}.
    \end{array}
    \end{align*}
    where $c_1, c_2, c_3$ are clauses. \hfill $\lhd$
\end{Example}
An aggregate atom is of the form:
\begin{align}\label{eq:aggregate}
    \textsc{sum}\{w_1 : l_1, \ldots, w_n : l_n\} \geq b
\end{align}
where $n \geq 1$, $l_1,\ldots,l_n$ are distinct literals, and $b,w_1,\ldots,w_n$ are positive integers.
For an atom $p$ of the form (\ref{eq:aggregate}), $\mathit{elem}(p) := \{(w_i,l_i) | i \in [1..n]\}$, $\mathit{lits}(p) := \{l | (w, l) \in \mathit{elem}(p) \}$, and $\mathit{bound}(p) := b$.
Moreover, $\textsc{count}\{l_1, \ldots, l_n\} \geq b$ denotes a shortcut for $\textsc{sum}\{w_1: l_1, w_2: l_2, \ldots, w_n: l_n\} \geq b$ where $w_1 = w_2 = \ldots = w_n = 1$.

An ASP \emph{program} $\Pi$ is a finite set of rules of the following form:
\begin{align}
    \label{eq:rule}
    p_1 \vee \cdots \vee p_m \leftarrow{} & \lnot p_{m+1}, \ldots, \lnot p_{k}, p_{k+1}, \ldots, p_n
\end{align}
where $n > 0$ and $n \geq m$, $p_1,\ldots,p_m$ are atoms, $p_{m+1}, \ldots, p_n$ are atoms or aggregate atoms.
For a rule $r$ of the form (\ref{eq:rule}), let $H(r)$ denote the set $\{p_1,\ldots,p_m\}$ of head atoms, and $B(r)$ denote the set $\{\lnot p_{m+1},\ldots, \lnot p_k, p_{k+1}, \ldots,p_n\}$ of body literals.
A rule $r$ of the form (\ref{eq:rule}) is said to be \textit{disjunctive} if $m \geq 2$, \textit{normal} if $m=1$, and a \textit{constraint} if $m=0$.
Moreover, a rule $r$ of the form $\{p_1, p_2, \ldots, p_m\} \leftarrow{} \lnot p_{m+1}, \ldots, \lnot p_{k}, p_{k+1}, \ldots, p_n$ is called \emph{choice rule} and defined here as a shortcut for the following rules: $p_1 \vee p_1' \leftarrow{} B(r),$
$p_2 \vee p_2' \leftarrow{} B(r),$
$\ldots,$
$p_m \vee p_m' \leftarrow{} B(r),$
where $p_1', \ldots, p_m'$ are fresh atoms not appearing in other rules.
Note that modern ASP solvers do not usually create any auxiliary atom to handle choice rules.
For an expression (SAT formula or ASP program) $\gamma$, $\mathit{atoms}(\gamma)$ denotes the set of (aggregate) atoms occurring in $\gamma$.

\begin{Example}\label{ex:asp}
    Let $\Pi_\mathit{run}$ be the following program:
    \begin{align*}
    \begin{array}{ll}
    r_1:& p_1 \vee p_4 \leftarrow{} \\
    r_2:& p_2 \vee p_3 \leftarrow{} \lnot p_4, p_1 \\
    r_3:& p_5 \leftarrow{} \textsc{sum}\{1 : p_1, 2 : p_2, 4 : p_4\} \geq 7.
    \end{array}
    \end{align*}
    Note that $r_1$ and $r_2$ are disjunctive rules, $r_3$ is a normal rule, and $\textsc{sum}\{1 : p_1, 2 : p_2, 4 : p_4\} \geq 7$ is an aggregate atom.
    \hfill $\lhd$
\end{Example}

The \emph{dependency graph} $G_\Pi$ of $\Pi$ has nodes $\mathit{atoms}(\Pi)$, and an arc $xy$, where $x$ and $y$ are (aggregate) atoms, for each rule $r \in \Pi$ such that $x \in H(r)$ and $y \in B(r)$.
An atom is \emph{recursive} in $\Pi$ if it is involved in a cycle of $G_\Pi$.
In the following, every program $\Pi$ is assumed to have no recursive aggregate atoms.
Note that $\Pi_\mathit{run}$ has such a property.


\paragraph{Semantics.}
An interpretation $I$ is a set of (aggregate) atoms.
Given an interpretation $I$, relation $\models$ is defined as follows:
\begin{itemize}
    \item for an atom $p$, $I \models p$ if $p \in I$; while $I \models \lnot p$ if $p\not\in I$;
    \item for an aggregate atom $p$ of the form (\ref{eq:aggregate}), $I \models p$ if $\sum_{(w,l) \in \mathit{elem}(p),\ I \models l}{w} \geq \mathit{bound}(p)$;
    while $I \models\lnot p$ if $\sum_{(w,l) \in \mathit{elem}(p),\ I \models l}{w} < \mathit{bound}(p)$;
    \item for a clause $c$, $I \models c$ if $I \models l$ for some $l \in c$;
    \item for a rule $r$ of the form (\ref{eq:rule}), $I \models B(r)$ if $I \models l$ for all $l \in B(r)$, $I \models H(r)$ if $I \models p$ for some $p \in H(r)$, and $I \models r$ if $I \models H(r)$ whenever $I \models B(r)$;
    \item for a SAT formula $\varphi$, $I \models \varphi$ if $I \models c$ for all $c \in \varphi$;
    \item for a program $\Pi$, $I \models \Pi$ if $I \models r$ for all $r \in \Pi$.
\end{itemize}
For an expression (SAT formula or ASP program) $\gamma$, $I$ is a model of $\gamma$ if $I \models \gamma$.

\begin{Example}
Consider $\varphi_{run}$ of Example~\ref{ex:cnf} and $\Pi_{run}$ of Example~\ref{ex:asp}, and $I=\{p_4\}$. $I \models \varphi_{run}$ and $I \models \Pi_{run}$.
\hfill $\lhd$
\end{Example}

The \emph{reduct} $\Pi^I$ of a program $\Pi$ with respect to an interpretation $I$ is $\{H(r) \leftarrow{} B(r) \ :\  r \in \Pi, I \models B(r)\}$~\cite{DBLP:journals/ai/FaberPL11}.
A model $I$ is a \emph{stable model} of a program $\Pi$ if there is no model $J$ of $\Pi^I$ such that $J \subset I$.

\begin{Example}
Consider $\Pi_{run}$ of Example~\ref{ex:asp} and $I=\{p_4\}$.
The reduct $\Pi_{run}^{I}$ is equal to $p_1 \vee p_4 \leftarrow{}$. Thus, $I$ is a stable model.
\hfill $\lhd$
\end{Example}

\subsection{Input format}
Modern SAT and ASP solvers usually take as input CNF formulae and ASP programs represented by means of a numeric format. 
Concerning SAT, the numeric format is called \emph{DIMACS}. Figure \ref{fig:dimacscnf} shows the representation of the formula $\varphi_{run}$ of Example~\ref{ex:cnf} in the DIMACS format.
\begin{figure}[t]
\figrule
$$
\begin{array}{l}
\texttt{p cnf 5 3}\\
\texttt{1 -3 0}\\
\texttt{2 3 -1 -4 0}\\
\texttt{-5 -4 0}\\
\end{array}
$$
\figrule
\caption{SAT formula $\varphi_{run}$ encoded in the DIMACS format.}\label{fig:dimacscnf}
\end{figure}
The first line, starting by \textit{p}, gives information about the formula: the instance is in CNF, and the numbers of atoms and clauses, respectively, are provided. In the DIMACS format each atom is uniquely identified by a number.
After the initial descriptive line, clauses are listed. Each clause is a sequence of distinct non-null numbers ending with \texttt{0} on the same line. Positive numbers denote the corresponding positive literals, while negative numbers represent negative literals. 

Concerning ASP programs, they are usually represented in the \textit{lparse format} \cite{lparse}. 
As in the DIMACS format, each atom is uniquely identified by a number and each rule is represented by a sequence of numbers.
Rules are listed, and each rule starts with an identifier of the rule type, as follows:
\begin{itemize}
    \item $1$ represents normal rules and constraints;
    \item $2$ represents aggregate atoms of the type \textsc{count};
    \item $3$ represents choice rules;
    \item $5$ represents aggregate atoms of the type \textsc{sum};
    \item $8$ represents disjunctive rules.
\end{itemize}

Figure \ref{fig:lparseasp} shows the representation of the program $\Pi_{run}$ of Example~\ref{ex:asp} in the lparse format.
\begin{figure}[t]
\figrule
$$
\begin{array}{l}
\texttt{8 2 2 3 0 0}\\
\texttt{8 2 4 5 2 1 2 3}\\
\texttt{5 6 7 3 0 3 5 2 1 2 4}\\
\texttt{1 7 1 0 6}\\
\vdots
\end{array}
$$
\figrule
\caption{ASP program $\Pi_{run}$ encoded in the lparse format.}\label{fig:lparseasp}
\end{figure}
In particular, consider line \texttt{8 2 4 5 2 1 2 3} representing the disjunctive rule $r_2$. The first number, \texttt{8}, is the identifier of the rule type, the number \texttt{2} represents $|H(r_2)|$, and \texttt{4} and \texttt{5} are the numeric identifiers of atoms $p_2$ and $p_3$, respectively.
Then, \texttt{2} and \texttt{1} represent $|B(r_2)|$ and $|B(r_2)^-|$, respectively. Finally, \texttt{2} and \texttt{3} are the identifiers of the atoms $p_4$ and $p_1$, respectively.
Concerning the aggregate atom appearing in rule $r_3$, this is represented by the line \texttt{5 6 7 3 0 3 5 2 1 2 4}, where \texttt{5} is the identifier of the rule type, the number \texttt{6} represents the numeric identifier of the aggregate atom, \texttt{7} is the bound of the aggregate atom, then \texttt{3} and \texttt{0} represent the number
of literals and negative literals in $\mathit{elem}()$, respectively. Then, literals $p_1$, $p_2$ and $p_4$ are listed followed by their corresponding weights.
Note that rule $r_3$ is represented by \texttt{1 7 1 0 6}, where \texttt{7} and \texttt{6} are the identifiers of $p_5$ and of the aggregate atom, respectively.

\subsection{Automated Configuration Techniques}
Many algorithms have parameters that can be adjusted to optimise performance (in terms of, e.g. solution cost, or runtime to solve
a set of instances). 
Formally, this problem can be stated as follows: given a
parameterised algorithm with possible configurations $\mathcal{C}$,
a benchmark set $\Pi$, and a performance metric $m(c,\pi)$ that measures the performance of a configuration $c\in\mathcal{C}$ on an instance
$\pi\in\Pi$ (the lower the better), 
find a configuration $c\in\mathcal{C}$ that minimises
$m$ over $\Pi$, i.e., that minimises
\begin{equation}
\label{eq:ac}f(c) = \frac{1}{|\Pi|} \sum_{\pi\in\Pi} m(c,\pi).
\end{equation}

The AI community has developed dedicated algorithm configuration systems to tackle this problem~\cite
  {DBLP:journals/jair/HutterHLS09,DBLP:conf/cp/AnsoteguiST09,DBLP:conf/ieaaie/YuanSB10}. In this work we exploit the sequential model-based algorithm configuration method SMAC~\cite{DBLP:conf/lion/HutterHL11}, which represents the state of the art of configuration tools and, differently from other existing tools, can handle continuous parameters. 
SMAC uses predictive models of algorithm performance~\cite{DBLP:journals/ai/HutterXHL14}
to guide its search for good configurations. It uses
previously observed $\langle{}$configuration, performance$\rangle{}$ pairs $\langle{}c, f(c)\rangle{}$ and
supervised machine learning (in particular,
random forests~\cite{DBLP:journals/ml/Breiman01})
to learn a function $\hat{f}:\mathcal{C} \rightarrow \mathds{R}$ that predicts the
performance of arbitrary parameter configurations,
and is used to select a promising configuration. Random forests are collections of regression trees, which are similar to decision trees but have real values (here: CPU-time performance) rather than class labels at their leaves. Regression trees are known to perform well for categorical input data. Random forests share this benefit and typically yield more accurate predictions; they also allow to quantify the uncertainty of a prediction. 
The performance data to fit the predictive models are collected sequentially. 

In a nutshell, after an
initialisation phase, SMAC iterates the following
three steps: (1)~use the performance measurements observed so far to
fit a random forest model $\hat{f}$; (2)~use $\hat{f}$ to select a
promising configuration $c \in \mathcal{C}$ to evaluate next,
trading off exploration of new parts of the configuration space
and exploitation of parts of the space known to perform well; and
(3)~run $c$ on one or more benchmark instances and compare its
performance to the best configuration observed so far. 

In order to save time in evaluating new configurations, SMAC first evaluates them on a single training instance; additional evaluations are only carried out (using a doubling schedule) if, based on the evaluations to date, the new configuration appears to outperform SMAC's best known configuration.  
Once the same number of runs has been evaluated for both configurations, if the new configuration still performs better then SMAC updates its best known configuration accordingly.

SMAC is an \emph{anytime algorithm} (or \emph{interruptible algorithm}) that interleaves the exploration of new configurations with additional runs of the current best configuration to yield both better and more confident results over time. As all anytime algorithms, SMAC improves performance over time, and for finite configuration spaces it is guaranteed to converge to the optimal configuration in the limit of infinite time.

SMAC has been used for configuring knowledge models in the fields of AI Planning \cite{DBLP:conf/ijcai/VallatiHCM15,DBLP:conf/aips/VallatiS18} and Abstract argumentation \cite{DBLP:journals/ijar/CeruttiVG18}.

\section{Knowledge Configuration}
\label{sec:know}


In both SAT formulae and ASP programs, clauses and rules are usually not ordered following a principled approach, but they are ordered according to the way in which the randomised generator has been coded, or  following the way in which information from the application domain has been collected, or deliberately shuffled to prevent potential biases. This is also generally true for the order in which literals of a given clause are presented in the formula, or in the program, with some differences among SAT clauses and ASP rules. However, we should consider that rules must preserve the head-body structure, so only rule's bodies are amenable to configuration; moreover,  literals in the positive and negative parts of the body can not be mixed. On the other hand, ASP programs contain further degrees of freedom given that the ASP language allows for a number of additional constructs, like aggregates.  

In this section we focus on the following question: \textit{given the set of clauses/rules, and the set of corresponding literals, in which order should they be listed to maximise the performance of a given solver, taking into account for ASP existing constraints above-mentioned and more constructs?} The underlying hypothesis is that the
order in which clauses, rules and literals are listed can be tuned to highlight elements that are important for satisfying, or demonstrating the unsatisfability, of the considered instance by the considered solver. 
%
%
To answer the above question, here we explain how we have configured SAT formulae and ASP programs, i.e., what features have been considered, and how related scores have been computed. 
Noteworthy, there is a significant body of work in both SAT and ASP that deal with features selection and computation, and they are outlined in Section \ref{sec:rel}; they are mainly concerned at instance-level, while our goal is to analyse the structure also at clause/rule level, for ordering among those elements. 

\subsection{Configuration of SAT Formulae}




The CNF configuration has to be performed online: as soon as a new formula is provided as input, the formula has to be configured before being presented to the solver. In a nutshell, given a set of parameters that can be used to modify the ordering of some aspect of the CNF formula, and given the value assigned to each parameter, the online configuration is performed by re-ordering clauses and literals accordingly. Notably, the value of each parameter has to be provided, and can be identified via an appropriate off-line learning step.

Given the depicted online scenario, 
we are restricted to information about the CNF that can be quickly gathered and that are computationally cheap to extract. Furthermore, the configuration must consider only general aspects that are common to any CNF. As it is apparent, the use of a computationally expensive configuration of a single CNF, that considers elements that are specific to the given CNF, would nullify the potential performance improvement, by drastically reducing the time available for the solver to find a solution (or to demonstrate unsatisfiability). 

In this work, we consider the possibility to order \textit{clauses} according to the following criteria, denoted as $\mathcal{F}_c$:
\begin{enumerate}[(c1)]
    \item the number of literals of the clause ($\mathit{size}$); \label{item:sat:clauselength}
    \item the fact that the clause is binary ($\mathit{bin}$); \label{item:sat:bin}
    \item the fact that the clause is ternary ($\mathit{ter}$); \label{item:sat:ter}
    \item the number of positive literals of the clause ($\mathit{positive}$); \label{item:sat:poslits}
    \item the number of negative literals of the clause ($\mathit{negative}$); \label{item:sat:neglits}
    \item the fact that the clause is binary, and both literals are negative ($\mathit{bin\_neg}$);
    \item the fact that the clause has only one negative literal ($\mathit{only\_one\_neg}$).\label{item:sat:oneneg}
\end{enumerate}

\textit{Atoms} can be listed in clauses according to the following criteria, denoted as $\mathcal{F}_m$:
\begin{enumerate}[(m1)]
    \item the number of clauses in which the atom appears ($\mathit{occ}$); \label{item:sat:occ}
    \item the average size of the clauses in which the atom is involved ($\mathit{occ\_avg}$); \label{item:sat:averagesize}
    \item the number of binary clauses in which the atom in involved ($\mathit{occ\_bin}$); \label{item:sat:occbin}
    \item the number of ternary clauses in which the atom is involved ($\mathit{occ\_ter}$); \label{item:sat:occter} 
    \item the number of times the atom appears in clauses as positive ($\mathit{occ\_pos}$);  \label{item:sat:occpos}
    \item the number of times the atom appears in clauses as negative ($\mathit{occ\_neg}$); \label{item:sat:occneg}
    \item the number of times the atom is involved in clauses where all literals are positive ($\mathit{occ\_all\_pos}$); \label{item:sat:occallpos} 
    \item the number of times the atom is involved in clauses where all literals are negative ($\mathit{occ\_all\_neg}$). \label{item:sat:occallneg}
\end{enumerate}

Moreover, we also include two additional categorical selectors, denoted as $\mathcal{F}_s$:
\begin{enumerate}[(s1)]
    \item to enable/disable the ordering of literals in the clauses ($\mathit{ord\_lit}$);
    \item to order clauses according to the ordering (direct or inverse) followed by the involved literals ($\mathit{ord\_cl}$);
\end{enumerate}
The set of proposed ordering criteria, denoted as $\mathcal{F}= \mathcal{F}_c \cup \mathcal{F}_m \cup \mathcal{F}_s$, is aimed at being as inclusive as possible, so that different characterising aspects of clauses and atoms can be taken into account, at the same time, for the configuration process. 

It is easy to notice that many of the introduced criteria focus on aspects of binary and ternary clauses. This is due to their importance in the search process. For instance, binary clauses are responsible, to a great degree, of unit propagation. There are also criteria that aim at identifying potentially relevant aspects. For instance, criterion (c\ref{item:sat:oneneg}) aims at identifying clauses that may represent implication relations between literals.    

There are different ways for encoding the identified degrees of freedom in CNFs as parameters. This is due to the fact that orders are not natively supported by general configuration techniques \cite{DBLP:conf/lion/HutterHL11,DBLP:conf/ecai/KadiogluMST10}. Results presented by  \cite{DBLP:conf/ijcai/VallatiHCM15} suggest that purely categorical parametrisations are not indicated for the configuration of models, as they tend to fragment the configuration space and to introduce discontinuities. Those combined aspects make the exploration of the configuration space particularly challenging for learning approaches. For this reason, here we generate $7$ continuous parameters for configuring the order of clauses, and $8$ continuous parameters for configuring the order of variables in clauses. Each parameter corresponds to one of the aforementioned criteria, and they have to be combined to generate different possible orderings of clauses and literals in CNFs. Each continuous parameter in $\mathcal{F}_c$ and $\mathcal{F}_m$ has associated a real value in the interval $[-10.0, +10.0]$ which represents (in absolute value) the \emph{weight} given to the corresponding ordering criterion.
Concerning selectors, $\mathit{ord\_lit}$ can assume a Boolean value, 0 or 1, whereas $\mathit{ord\_cl}$ can be 0 if clauses must be ordered using only the features of the literals appearing in the clause, 1 if clauses must be ordered using only the features of the clause, 2 if clauses must be ordered using both the features of the literals and the features of the clauses.
%
%
Thus, the configuration space is $\mathcal{C} = [-10.0,+10.0]^{15} \times \{0, 1\} \times \{0, 1, 2\}$.
A (total) function $\omega : \mathcal{F} \mapsto [-10.0,+10.0]$ maps parameters in $\mathcal{F}$ to a weight, where $\omega(\mathit{ord\_lit})$ is restricted to be in $\{0,1\}$ and $\omega(\mathit{ord\_cl})$ is restricted to be in $\{0,1,2\}$, respectively.

The configuration criteria mentioned above can be used to order any CNF.
In particular, given a CNF $\varphi$ and a weight function $\omega$, the corresponding configuration of the formula is obtained as follows. For each atom $p$ occurring in $\varphi$, an ordering score of $p$ is defined as:

\begin{equation}
\label{eq:ordering}O_{at}(p, \varphi, \omega) =  \sum_{c \in \mathcal{F}_m} (value(p, \varphi, c) \cdot \omega(c))
\end{equation}  

where $c$ is a criterion for configuring literals' order in the set $\mathcal{F}_m$ (i.e., from (m\ref{item:sat:occ}) to (m\ref{item:sat:occallneg})), and $value(p, \varphi, c)$ is the numerical value of the corresponding aspect for the atom $p$.
If $\omega(\mathit{ord\_lit}) = 1$, then, for every clause, the involved literals are ordered (in descending order) following the score $O_{at}$. Ties are broken following the order in the original CNF configuration. As it is apparent from equation (\ref{eq:ordering}), a positive (negative) value of $\omega(c)$ can be used to indicate that the aspect corresponding to the parameter $c$ is important for the SAT solver, and that literals with that aspect should be listed early (late) in the clause to improve performance.
If $\omega(\mathit{ord\_lit}) = 0$, literals follow the order as in the provided initial CNF.

Similarly to what is presented in equation (\ref{eq:ordering}) for literals, clauses are ordered according to a corresponding score $O_{cl}(\mathit{cl}, \varphi, \omega)$, defined as follows:
\begin{equation}\label{eq:ordering2}
    O_{cl}(\mathit{cl}, \varphi, \omega)=
    \begin{cases}
      \sum_{p \in \mathit{cl}^+ \cup \overline{\mathit{cl}^-}} O_{at}(p, \varphi, \omega) & \text{if}\ \omega(\mathit{ord\_cl})=0 \\
      \sum_{c \in \mathcal{F}_c} (\mathit{value}(\mathit{cl}, \varphi, c) \cdot \omega(c))  & \text{if}\ \omega(\mathit{ord\_cl})=1\\
      \sum_{p \in \mathit{cl}^+ \cup \overline{\mathit{cl}^-}} O_{at}(p, \varphi, \omega) +  \sum_{c \in \mathcal{F}_c} \mathit{value}(\mathit{cl}, \varphi, c) & \text{if}\ \omega(\mathit{ord\_cl})=2
    \end{cases}
\end{equation}

\begin{figure}[t]
\figrule
$$
\begin{array}{llll}
& \texttt{p cnf 5 3} & & \texttt{p cnf 5 3}\\
\varphi_{\mathit{run:}}& \texttt{1 -3 0} & \qquad\varphi_{\mathit{conf:}}& \texttt{3 -1 -4 2 0}\\
& \texttt{2 3 -1 -4 0} & & \texttt{-4 -5 0}\\
& \texttt{-5 -4 0} & &\texttt{1 -3 0}\\
\end{array}
$$
\figrule
\caption{The example CNF formula non configured ($\varphi_{run}$), and the configured version ($\varphi_{\mathit{conf}}$). Configuration has been done by listing clauses according to their length and the number of negative literals. Literals are listed following the number of clauses they are involved.}\label{fig:conf-sat}
\end{figure}

\begin{Example}
Let us consider again the CNF $\varphi_{run}$ of Example~\ref{ex:cnf} and reported, using the DIMACS format, in Figure \ref{fig:conf-sat}. Suppose that we are interested in listing clauses according to their length (criterion (c\ref{item:sat:clauselength})) and to the number of involved negative literals (criterion (c\ref{item:sat:neglits})). Similarly, we are interested in listing the literals of a clause according to the number of clauses in which they appear (criterion (m\ref{item:sat:occ})).
In this case, we have to set $\omega(\mathit{size}) = 10.0$, $\omega(\mathit{negative}) = 10.0$, $\omega(\mathit{occ}) = 10.0$, $\omega(\mathit{ord\_lit}) = 1$, and $\omega(\mathit{ord\_cl}) = 1$, whereas $\omega(c) = 0.0$ for all other criterion $c$ in $\mathcal{F}$.
Then, $O_{\mathit{cl}}("\texttt{2 3 -1 -4 0}", \varphi_{run}, \omega) = 60.0$, since it involves 4 literals, and 2 of them are negative, thus $4 \cdot 10.0 + 2 \cdot 10.0 = 60.0$.
According to the same criteria, $O_{\mathit{cl}}("\texttt{1 -3 0}", \varphi_{run}, \omega) = 30.0$.
In a similar way, but considering the corresponding criterion, the score of literals can be calculated, and literals are then ordered accordingly in each clause.
Result is $\varphi_{\mathit{conf}}$ reported in Figure \ref{fig:conf-sat}.
Note that the first line of the considered CNF formula is unmodified, as the DIMACS format require it to be the first, and to present information in a given order. \hfill $\lhd$
\end{Example}

\vspace{0.3cm}

The way in which the considered ordering criteria are combined, via equations (\ref{eq:ordering}) and (\ref{eq:ordering2}), gives a high degree of freedom for encoding and testing different configurations. Very specific aspects can be prioritised: for instance, it would be possible to present first clauses that are binary, and where both literals are positive, by penalising criterion (c\ref{item:sat:neglits}) and giving a high positive weight to criterion (c\ref{item:sat:bin}). Furthermore, additional criteria can be added, with no need to modify or update the overall configuration framework.

\subsection{Configuration of ASP Programs}\label{sec:configasp}






In this subsection, instead, we turn our attention to the configuration of ASP programs.
Similarly to SAT, we generate \ref{item:end} continuous parameters for configuring the order of rules and aggregates. Each parameter corresponds to a feature that is syntactic and easy to compute, and they have  to  be  combined  to  generate  different  possible  orderings  of  rules and aggregates.
Each  continuous  parameter has an associated real value in the interval $[-10.0, +10.0]$ which represents (in absolute value) the \textit{weight} given to the corresponding ordering criterion.
The continuous parameters are detailed in the following:
\begin{enumerate}[(k1)]
    \item occurrences of a literal in heads ($\mathit{head\_occ}$) \label{item:initocc}
    \item occurrences of a literal in bodies ($\mathit{body\_occ}$)
    \item occurrences of a literal in positive part of bodies ($\mathit{pos\_body\_occ}$)
    \item occurrences of a literal in negative part of bodies ($\mathit{neg\_body\_occ}$)
    \item occurrences of a literal in bodies of "short" size ($\mathit{short\_body\_occ}$)
    \item occurrences of a literal in positive part of bodies of "short" size (\textit{short\_pos\_body\_occ})
    \item occurrences of a literal in negative part of bodies of "short" size (\textit{short\_neg\_body\_occ})  
    \item occurrences of a literal in aggregates ($\mathit{aggregate\_occ}$) \\ \label{item:endocc}
    \item constraints ($\mathit{constraints}$)
    \item normal rules ($\mathit{normal}$)
    \item disjunctive rules ($\mathit{disjunctive}$)
    \item choice rules  ($\mathit{choice}$)
    \item literals in the body ($\mathit{body}$)\label{item:initothers}
    \item literals in the positive part of the body ($\mathit{p\_body}$)
    \item literals in the negative body of the body ($\mathit{n\_body}$)
    \item ratio between positive and negative body literals ($\mathit{ratio\_pos\_neg}$)
    \item Horn bodies ($\mathit{horn}$)\label{item:horn}
    \item recursive atoms in heads ($\mathit{rec\_head}$)
    \item recursive atoms in bodies ($\mathit{rec\_body}$)
    \item binary or ternary rules ($\mathit{short}$)\\\label{item:endothers}
    \item aggregates ($\mathit{aggregate}$) \label{item:aggr}
    \item aggregate size ($\mathit{aggregate\_size}$) \label{item:aggrsize}
    \item ratio between aggregate size and bound ($\mathit{aggregate\_ratio\_bound\_size}$) \label{item:end}
\end{enumerate}

\begin{function}[t]
    s := 0\;
    s += $|\{r \ :\  r \in \Pi, l \in H(r)\}| * \omega(\mathit{head\_occ})$\; \label{ln:init_occ}
    s += $|\{r \ :\  r \in \Pi, l \in B(r)\}| * \omega(\mathit{body\_occ})$\;
    s += $|\{r \ :\  r \in \Pi, l \in B(r)^+\}| * \omega(\mathit{pos\_body\_occ})$\; \label{ln:posbodyocc}
    s += $|\{r \ :\  r \in \Pi, l \in B(r)^-\}| * \omega(\mathit{neg\_body\_occ})$\;
    s += $|\{r \ :\  r \in \Pi, l \in B(r), |B(r)| \leq 2\}| * \omega(\mathit{short\_body\_occ})$\;
    s += $|\{r \ :\  r \in \Pi, l \in B(r)^+, |B(r)| \leq 2\}| * \omega(\mathit{short\_pos\_body\_occ})$\;
    s += $|\{r \ :\  r \in \Pi, l \in B(r)^-, |B(r)| \leq 2\}| * \omega(\mathit{short\_neg\_body\_occ})$\;
    s += $|\{p \ :\ p \in \mathit{atoms}(\Pi), \mathit{p\ is\ an\ aggregate\ atom}, l \in \mathit{lits}(p)\}| * \omega(\mathit{aggregate\_occ})$\; \label{ln:end_occ}
    \Return s;
    \caption{$O_{l}$(Literal $l$, Program $\Pi$, Weight function $\omega$)}\label{fn:occ}
\end{function}
\begin{function}[t]
    s := 0\;
    \lIf{$|H(r)| = 0$}{ s += $\omega(constraint)$}\label{ln:initor}
    \lIf{$|H(r)| = 1$}{ s += $\omega(normal)$}
    \lIf{$|H(r)| > 1$}{ s += $\omega(disjunctive)$}
    \lIf{$r$ is choice}{ s += $\omega(choice)*t_1$}\label{ln:endor}
    
    s += $|B(r)| * \omega(body)$ + $|B(r)^+| * \omega(p\_body)$ + $|B(r)^-| * \omega(\mathit{n\_body})$\; \label{ln:body}
    \lIf{$|B(r)^-| \neq 0$} { s += ($|B(r)^+| \div |B(r)^-|$) $*$ $\omega(\mathit{ratio\_pos\_neg})$ }\label{ln:ratio}
    \lIf{$|B(r)^+| = 1$} { s += $\omega(\mathit{horn})$ }\label{ln:horn}
    s += $|\{p \in H(r) |\ p\ is\ recursive\}| * \omega(\mathit{rec\_head})$\; \label{ln:rechead}
    s += $|\{l \in B(r) |\ l\ is\ recursive\}| * \omega(\mathit{rec\_body})$\; \label{ln:recbody}
    \lIf{$|H(r)|+|B(r)| \geq 2$ \textbf{and} $|H(r)|+|B(r)| \leq 3$ \label{ln:short}} { s+= $\omega(\mathit{short})$}
    s += $(\sum_{l \in H(r) \cup B(r)}$ \ref{fn:occ}($\Pi$, $l$, $\omega$)) $\div\ (|H(r)| + |B(r)|)$\;\label{ln:sumocc}
    \Return s;
    \caption{$O_r$(Rule $r$, Program $\Pi$, Weight function $\omega$)}\label{fn:rule}
\end{function}
\begin{function}[t]
    s := $t_2 * \omega(aggregate) + |\mathit{lits(p)}| * \omega(\mathit{aggregate\_size})$\; \label{ln:initaggr}
    s += $(bound(p) \div \sum_{(w, l) \in \mathit{elem}(p)} w) * \omega(\mathit{aggregate\_ratio\_bound\_size})$\; \label{ln:boundsize}
    s += $(\sum_{l \in \mathit{lits}(p)}$ \ref{fn:occ}($\Pi$, $l$, $\omega$)) $\div$ $|\mathit{lits(p)}|$\; \label{ln:occaggr}
    \Return s;
    \caption{$O_a$(Aggregate atom $p$, Program $\Pi$, Weight function $\omega$)}\label{fn:aggr}
\end{function}

Given the structure of ASP programs, richer than SAT formulae, it is not as straightforward and compact to calculate scores as for SAT formulae; in order to calculate the final score, we have introduced three functions \ref{fn:occ}, \ref{fn:rule} and \ref{fn:aggr} for calculating scores for literals, rules and aggregate atoms, respectively, that take as input a program, an element (a literal, a rule or an aggregate atom, respectively), and a weight function $\omega : \mathcal{F} \mapsto [-10.0, 10.0]$, where $\mathcal{F} = \{\mathit{head\_occ}, \mathit{body\_occ}, \ldots, \mathit{aggregate\_ratio\_bound\_size}\}$, i.e., it includes all the features reported from (k\ref{item:initocc}) to (k\ref{item:end}). The output of the three functions is the score of the element, computed as a sum of individual contributions brought by the features linked to the element.
The score of rules and aggregates is later on used to order the rules of ASP programs. 

\paragraph{Function \ref{fn:occ}} computes the score of a given literal $l$ by summing up, from line \ref{ln:init_occ} to \ref{ln:end_occ}, all single contributions of features (k\ref{item:initocc})--(k\ref{item:endocc}), by multiplying the number of times $l$ "falls" in the category described by the respective feature to the weight of the feature.
As an example, line \ref{ln:posbodyocc}, related to feature $\mathit{pos\_body\_occ}$, gives a contribution to $s$ obtained by multiplying the number of times literal $l$ occurs in positive bodies of the program $\Pi$ and the weight of the feature.

\paragraph{Function \ref{fn:rule}} computes the score of a rule $r$.
Depending of whether $r$ is a constraint, a normal or disjunctive, or a choice rule, one of the lines from \ref{ln:initor} to \ref{ln:endor} is activated. If $r$ is a choice, an additional factor $t_1$ is considered, which is an arbitrary large value, set to $10^5$ in our experiments, and (from the configuration side) means to put priorities to such rules.
Then, lines from \ref{ln:body} to \ref{ln:sumocc} contribute further to the score, as a bonus, for features (k\ref{item:initothers})--(k\ref{item:endothers}): lines \ref{ln:body}, \ref{ln:rechead}, and \ref{ln:recbody} work similarly as within function \ref{fn:occ}, while lines \ref{ln:ratio}, \ref{ln:horn}, and \ref{ln:short} behave similarly to lines \ref{ln:initor}-\ref{ln:endor} in this function for the respective feature.
Finally, line \ref{ln:sumocc} employs function \ref{fn:occ} to compute a score that is later on divided by the number of literals appearing in the rule.

\paragraph{Function \ref{fn:aggr}} computes the score for an aggregate atom $a$. In particular, line \ref{ln:initaggr} takes into account features (k\ref{item:aggr}) and (k\ref{item:aggrsize}), related to the presence of aggregates and its size, giving a high reward (value $t_2$ set to $10^5$) to the presence of aggregates as for choice rules before, line \ref{ln:boundsize} considers the ratio between bound and size of the aggregate, while line \ref{ln:occaggr} has a similar behaviour as of line \ref{ln:sumocc} of the function \ref{fn:rule}.

\begin{figure}[t]
\centering
$$
\begin{array}{llll}
& \texttt{8 2 2 3 0 0} & & \texttt{5 6 7 3 0 3 5 2 1 2 4} \\
\Pi_{\mathit{run}}:& \texttt{8 2 4 5 2 1 2 3} & \qquad \Pi_{\mathit{conf}}: & \texttt{8 2 2 3 0 0}\\
& \texttt{5 6 7 3 0 3 5 2 1 2 4} & & \texttt{8 2 4 5 2 1 2 3}\\
& \texttt{1 7 1 0 6} & & \texttt{1 7 1 0 6}\\
& \qquad \vdots & & \qquad \vdots\\
\end{array}
$$
\caption{The example ASP program non configured ($\Pi_{run}$), and the configured version ($\Pi_{\mathit{conf}}$). Configuration has been done by preferring aggregates and rules with literals occurring in many negative bodies.}
\label{fig:conf-asp}
\end{figure}

\begin{Example}
Consider again the program $\Pi_{run}$ of Example~\ref{ex:asp} and its lparse representation, reported in Figure~\ref{fig:conf-asp}.
Suppose that we are interested in ordering the program by giving a high priority to aggregates and then to give additional priorities to rules according to the atoms that occur in negative bodies.
This can be done by leaving all the parameters to the default value $0.0$, but $\omega(\mathit{aggregate})$ and $\omega(\mathit{p\_body})$ that are both set to 10.0.
In particular, the atom with id 2 occurs in the negative body of the second rule of $\Pi_{run}$, while other atoms do not occur in the negative body. Thus, \ref{fn:occ}(2, $\Pi_{run}$, $\omega$) returns 10.0, whereas \ref{fn:occ}($l$, $\Pi_{run}$, $\omega$) returns 0.0 for $l \in \{1,3,4,5,6,7\}$.
Then, \ref{fn:rule}("8 2 2 3 0 0", $\Pi_{run}$, $\omega$) and \ref{fn:rule}("8 2 4 5 2 1 2 3", $\Pi_{run}$, $\omega$) return 5.0 and 2.5, respectively, whereas \ref{fn:rule}("1 7 1 0 6", $\Pi_{run}$, $\omega$) returns 0.0, and \ref{fn:aggr}("5 6 7 3 0 3 5 2 1 2 4", $\Pi_{run}$, $\omega$) returns 100003.33.
Result is $\Pi_{\mathit{conf}}$ reported in Figure \ref{fig:conf-asp}. \hfill $\lhd$

\end{Example}

\section{Experimental Analysis}
\label{sec:exp}

This experimental analysis aims at evaluating the impact of the proposed automated approach for performing the configuration of knowledge models, on state-of-the-art domain-independent solvers' performance from SAT and ASP.
\subsection{Experimental Settings}

In this work we use the state-of-the-art SMAC~\cite{DBLP:conf/lion/HutterHL11} configuration approach for identifying a configuration of the knowledge model, that aims at improving the number of solved instances and the PAR10 performance of a given solver. PAR10 is the average runtime where unsolved instances count as $10 \times$ cutoff time. PAR10 is a metric commonly exploited in machine learning and algorithm configuration techniques, as it allows to consider coverage and runtime at the same time \cite{DBLP:journals/jair/EggenspergerLH19}.

For each solver, a benchmark-set specific configuration was generated using SMAC 2.08. A dedicated script, either in Python 2.7 or in C++, is used as a wrapper for extracting information from a knowledge model and, according to the parameters' values, reconfigure it and provide it as input for the solver. 

Experiments, on both SAT and ASP instances, were run on a machine executing Linux Ubuntu 4.4.0-104 and equipped with Intel Xeon 2.50 Ghz processors. Each SMAC configuration process, i.e., for each pair $\langle solver, benchmark\ set\rangle$, has been given a budget of 7 sequential CPU-time days, and run on a dedicated processor.

To compare performance, as mentioned we rely on the number of solved instances, the PAR10, and the IPC score. 
For a solver $\cal R$ and an instance $p$ to be solved, {\it Score}$({\cal R},p)$ is defined as:
\vspace{0.1cm}
\begin{eqnarray*}
\hspace{-0.3cm}
  {\textrm Score}({\cal R},p) =
  \left\{ 
  \begin{array}{ll} \vspace{0.15cm}
    0  \hspace{1.3cm} & \mbox{\small \it if $p$ is unsolved} \\ 
    \frac{1}{1+ \log_{10}(\frac{T_{p}({\cal R})}{T^*_{p}})} & \mbox{\small \it otherwise} \\ 
  \end{array}
\right.\\
\end{eqnarray*}

\vspace{0.1cm}
\noindent
where $T^*_{p}$ is the minimum amount of time required by any compared system to solve the instance, and $T_{p}({\cal R})$ denotes the CPU time required by $\cal R$ to
solve the instance $p$.  Higher values of the score indicate better
performance, 
where the best performing solver obtains a score equals to 1.

All the executables, benchmarks, instances, and generators used in the experiments are available at \url{https://www.mat.unical.it/~dodaro/research/aspsatconfig}.

\subsection{Configuration of SAT Formulae}

We selected 3 SAT solvers, based on their performance in recent SAT competitions and their widespread use: \textsc{cadical} version sc17 \cite{cadLing}, \textsc{glucose} 4.0 \cite{DBLP:conf/sat/AudemardLS13}, and \textsc{lingeling} version bbc \cite{cadLing}. 

In designing this part of the experimental analysis, we followed the Configurable SAT Solver Challenge (CSSC) \cite{DBLP:journals/ai/HutterLBBHL17}. The competition aimed at evaluating to which extent SAT solvers' performance can be improved by algorithm configuration for solving instances from a given class of benchmarks. In that, the CSSC goals are similar to the goals of this experimental analysis, i.e., assessing how performance can be improved via configuration, thus their experimental settings are deemed to be appropriate for our analysis. However, CSSC focused on the configuration of SAT solvers' behaviour by modifying exposed parameters of solvers. In this work we do not directly manipulate the behaviour of SAT solvers via exposed parameters, but we focus on the impact that the configuration of a CNF formula can have on solvers. 

Following CSSC settings, a cutoff of 5 CPU-time minutes, and a memory limit of 8 GB of RAM, has been set for each solver run on both training and testing instances. This is due to the fact that many solvers have runtime distributions with heavy tails \cite{DBLP:journals/jar/GomesSCK00}, and that practitioners often use many instances and relatively short runtimes to benchmark solvers for a new application domain \cite{DBLP:journals/ai/HutterLBBHL17}. There is also evidence that rankings of solvers in SAT competitions would remain similar if shorter runtimes are enforced \cite{DBLP:journals/amai/HutterHL10}. 

We chose benchmark sets from the CSSC 2014 edition \cite{DBLP:journals/ai/HutterLBBHL17}, and the benchmarks used in the Agile track of the 2016 SAT competition.\footnote{https://baldur.iti.kit.edu/sat-competition-2016/} These two competitions provide benchmarks that can highlight the importance of configuration (CSSC) even though a different type of configuration than the one considered in this paper and that include instances that have to be solved quickly (Agile). In particular, CSSC benchmarks can allow us to compare the impact of the proposed CNF configuration with regard to the solvers' configuration.

Selected CSSC 2014 benchmark sets include: Circuit Fuzz (Industrial track), 3cnf, K3 (Random SAT+UNSAT Track), and Queens and Low Autocorrelation Binary Sequence (Crafted track).\footnote{http://aclib.net/cssc2014/benchmarks.html} Benchmark sets were selected in order to cover most of the tracks considered in CSSC, and by checking that at least 20\% of the instances were solvable by considered solvers, when run on the default CNFs. 
Benchmarks were randomly divided into training and testing instances, aiming at having between 150-300 instances for testing purposes, and a similar amount of benchmarks for training. The size of each testing set is shown in Table \ref{tab:exp1}.

\begin{table}[t]
\renewcommand{\arraystretch}{1.1}{
\begin{tabular}{lrrrrrrr}
    & & \multicolumn{2}{c}{\textsc{cadical}} & \multicolumn{2}{c}{\textsc{glucose}} & \multicolumn{2}{c}{\textsc{lingeling}} \\
    \cline{3-4}\cline{5-6}\cline{7-8}
Problem &	\#	&	Def.	&	Conf.	&	Def.	&	Conf.	&	Def.	&	Conf.\\
\cline{1-8}
K3	&	150	&	61	&	\textbf{66}	&	78	&	\textbf{81}	&		74	&	\textbf{75}\\
3cnf	&	250	&	31	&	\textbf{34}	&	116	&	\textbf{119}	&		37	&	\textbf{40}\\
Queens	&	150	&	140	&	\textbf{141}		&	124	&	\textbf{125}	&	126	&	\textbf{127}\\
Low Autocorrelation	&	300	&	182	&	\textbf{184}	&	185	&	\textbf{191}	&		177	&	\textbf{180}\\
Circuit Fuzz	&	185	&	166	&	\textbf{168}	&		176	&	176	&		173	&	\textbf{175}\\
Agile16	&	250	&	219	&	\textbf{221}	&		226	&	\textbf{231}	&		195	&	\textbf{202}\\
\cline{1-8}
Total & 1285 & 799 & \textbf{814} & 905 & \textbf{923} & 782 & \textbf{799}\\
\cline{1-8}
\end{tabular}
}
\caption{\label{tab:exp1} Number of solved instances of the selected solvers on the considered benchmark set when running on the default and on the configured CNFs. Bold indicates the best result.}
\end{table}

Table \ref{tab:exp1} summarises the results of the selected SAT solvers on the considered benchmark sets. Results are presented in terms of the number of timeouts on testing instances, achieved by solvers run using either the default or the configured CNFs. Indeed, all of the considered solvers benefited from the configuration of the CNFs. Improvements vary according to the benchmark sets: the Agile16 set is, in general, the set where the solvers gained more by the use of configured CNFs. 
Remarkably, the improvements observed in Table \ref{tab:exp1} are comparable to those achieved in CSSC 2013 and 2014, that were achieved by configuring the solvers' behaviour \cite{DBLP:journals/ai/HutterLBBHL17}. In fact, these results may confirm our intuition that the way in which clauses and literals are ordered has an impact on the way in which solvers explore the search space. Listing ``important'' clauses earlier may lead the solver to tackle complex situations early in the search process, making it then easier to find a solution. In that, it may be argued that a solver's behaviour can be controlled internally, by modifying its exposed parameters, and externally by ordering the CNF in a suitable way.

Interestingly, the overall results (last row of Table \ref{tab:exp1}) indicate that the CNF configuration does not affect all the solvers in a similar way, and that can potentially lead to rank inversions in competitions or comparisons. This is the case of \textsc{lingeling} (on configured formulae) and \textsc{cadical} on "default" formulae. This may suggest that current competitions could benefit by exploiting a solver-specific configuration, in order to mitigate any implicit bias due to the particular CNF configuration exploited. Randomly listing clauses and variables may of course remove some bias, but it can also be the case that different biases are introduced. In that sense, allowing solvers to be provided with a specifically-configured CNF may lead to a better comparison of performance. Finally, it is worth noting that the way in which the CNFs are configured varies significantly between solvers, as well as according to the benchmark set. In other words, there is not a single ordering that allows to maximise the performance of all the SAT solvers at once.

\begin{table}[t]
{
\renewcommand{\arraystretch}{1.1}
\begin{tabular}{lrrrrrr}
    & \multicolumn{2}{c}{\textsc{cadical}} & \multicolumn{2}{c}{\textsc{glucose}} & \multicolumn{2}{c}{\textsc{lingeling}} \\
    \cline{2-3}\cline{4-5}\cline{6-7}
Problem &	Def.	&	Conf.	&	Def.	&	Conf.	&	Def.	&	Conf.\\
\cline{1-7}
K3 & 56.7 &  \textbf{59.9} &  71.3  &   \textbf{76.3} & 67.8  &   \textbf{68.6}\\
3cnf & 27.3 &  \textbf{31.6} & 106.6  &  \textbf{107.0} & 33.6  &  \textbf{35.9} \\
Queens & 136.5 & \textbf{137.6}  & 119.3  &  \textbf{121.1}  & 120.6  &  \textbf{122.9}  \\
Low Autocorrelation & 171.8  &  \textbf{173.4}  & 177.2  &  \textbf{183.7}  & 171.0  &  \textbf{175.3}  \\
Circuit Fuzz & 156.3  &  \textbf{160.8}  & 175.2  &  \textbf{175.3}  & 161.3  &  \textbf{164.3}  \\
Agile16 & 208.1  &  \textbf{211.3}  & 209.1  &  \textbf{215.9}  & 188.6  &  \textbf{196.6}  \\
\cline{1-7}
Total & 756.7  &  \textbf{774.6}  & 858.7  &  \textbf{879.3}  & 742.9  &  \textbf{763.6}  \\
\cline{1-7}
\end{tabular}
}
\caption{\label{tab:exp3} Results of the selected solvers on the considered benchmark sets. For each solver and benchmark, we show the IPC score achieved when running on the default and on the configured CNFs. Bold indicates the best result. Results of different solvers cannot be directly compared.}
\end{table}
\begin{table}[t]
{
\renewcommand{\arraystretch}{1.1}
\begin{tabular}{lrrrrrr}
    & \multicolumn{2}{c}{\textsc{cadical}} & \multicolumn{2}{c}{\textsc{glucose}} & \multicolumn{2}{c}{\textsc{lingeling}} \\
    \cline{2-3}\cline{4-5}\cline{6-7}
Problem &	Def.	&	Conf.	&	Def.	&	Conf.	&	Def.	&	Conf.\\
\cline{1-7}
K3 & 1788.9 & \textbf{1692.8} & 1448.7 & \textbf{1391.1}  & 1538.8 & \textbf{1521.5}\\
3cnf & 2640.7 & \textbf{2601.8} & 1660.5 &\textbf{1629.2}  & 2569.9 & \textbf{2534.7}\\
Queens & 217.2 & \textbf{196.3} & 526.3 & \textbf{507.6}  & 495.2 & \textbf{476.1}\\
Low Autocorrelation  & 1184.8 & \textbf{1166.3} & 1155.0 & \textbf{1099.6} & 1236.6 & \textbf{1208.9}\\
Circuit Fuzz & 324.7 & \textbf{294.1} & \textbf{159.4} & 159.5  & 216.1 & \textbf{199.3}\\
Agile16 & 417.7 & \textbf{391.0}  & 327.2 & \textbf{277.1} & 707.6 & \textbf{628.8}\\
\cline{1-7}
\end{tabular}
}
\caption{\label{tab:exppar10} Results of the selected solvers on the considered benchmark sets. For each solver and benchmark, we show the PAR10 score achieved when running on the default and on the configured CNFs. Bold indicates the best result.}
\end{table}


 
%


In Tables \ref{tab:exp3} and \ref{tab:exppar10} the performance of a solver run on the default and configured formulae are compared in terms of IPC score and PAR10. Results indicate that the configuration provides, for most of the benchmark sets, a noticeable improvement. 

To shed some light on the most relevant aspects of the SAT formula configuration, we assessed the importance of parameters in the considered configurations using the fANOVA tool \cite{DBLP:conf/icml/HutterHL14}. We observed that in most of the cases, improvements are mainly due to the effect of the correct configuration of a single criterion, rather than to the interaction of two or more criteria together. In terms of clauses, parameters controlling the weight of criteria (c\ref{item:sat:poslits}) and (c\ref{item:sat:neglits}) are deemed to be the most important: in other words, the number of positive (or negative) literals that are involved in a clause are a very important aspect for the performance of SAT solvers. The solver that can gain the most by ordering the clauses is \textsc{lingeling}. In particular, this solver shows best performance when clauses with a large number of negative literals are listed early. 

Parameters related to criteria (m\ref{item:sat:averagesize}), (m\ref{item:sat:occneg}), and (m\ref{item:sat:occallneg}) have shown to have a significant impact with regard to the literals' ordering in clauses. For \textsc{glucose} and \textsc{cadical}, criterion (m2), i.e., the average size of the clauses in which the literal is involved,     is the most important single criterion that has to be correctly configured. However, it is a bit hard to derive some general rules, as their impact on orderings vary significantly with regard to the solver and the benchmark set. 

Generally speaking, also in the light of the criteria that are most important for clauses, the ordering of literals appears to be the most important in a CNF formulae: this is also because, in many cases, clauses are ordered according to the (separately-calculated) weight of the involved literals. This behaviour can be due to the way in which data structures are generated by solvers: usually literals are the main element, that is also the focus of heuristic search used by SAT solvers. Instead, clauses from the CNF tend to have a less marked importance during the exploration of the search space, as they are related to literals mostly via lists, and are exploited only for checking satisfiability and performing unit propagation. Clauses learnt during the search process are not included in our analysis, as they are not part of the CNF formula--but are generated online by the solver.

Finally, we want to test if there is a single general configuration that improves the performance of a solver on any formula, despite of the benchmark and underlying structure. Therefore, we trained each of the considered solvers on a training set composed by an equal proportion of instances from each of the 6 benchmark sets. As for previous configurations, we gave 5 days of sequential CPU-time for each learning process, and obtained configurations have been tested on an independent testing set that includes instances from all the benchmark sets. Results are presented in Table \ref{tab:exp4}. 

\begin{table}[t]
{
\renewcommand{\arraystretch}{1.1}
    \begin{tabular}{lrrrr}
  
\hline
Solver & \multicolumn{2}{c}{Solved} & \multicolumn{2}{c}{IPC score}\\
\cline{2-3}\cline{4-5}
&  Def. & Conf. & Def. & Conf.\\
\hline
\textsc{cadical}  &  1184  &  \textbf{1187} & \textbf{172.7}  &  172.5\\
\textsc{glucose}  &  1205  &  \textbf{1207} & 207.5  &  \textbf{211.2}\\
\textsc{lingeling}  &  1176  &  \textbf{1177} & 190.7  &  \textbf{191.7}\\
\hline
\end{tabular}
}
\caption{\label{tab:exp4} Results achieved by the selected solvers on the general testing set. For each solver, we show the number of solved instances and IPC score achieved when running on the default and on the CNFs configured using the general configuration. Bold indicates the best result.}
\end{table}


Results on the independent testing set indicate that this sort of configuration has a very limited impact on solvers' performance. This seems to confirm our previous intuition that solvers require differently configured formulae according to the underlying structure of the benchmark: it is therefore the case that structurally different sets of instances require a very different configuration. Intuitively, this seems to point to the fact that, in different structures, the characteristics that identify challenging elements to deal with, vary. Solvers, when dealing with different sets of benchmarks, are then sensitive to different aspects of the CNF formulae, that should be appropriately highlighted and configured. On the one hand, this result may be not fully satisfying, as it suggests that there is not a quick way to improve the performance of SAT solvers. On the other hand, the results of the other experiments indicate that, for real-world applications of SAT where instances share some underlying structure, there is the possibility to further improve the SAT solving process by identifying a specific configuration for the solver at hand.
As a further observation, we remark that the presented results, achieved on testing set instances, are comparable to those observed on the training set used by SMAC for the configuration process. This confirms the ability of the learned knowledge to generalise on different instances.

\subsection{Configuration of ASP Programs}
We selected 3 ASP solvers, based on their performance in recent competitions, on the different approaches implemented, and for their widespread use: \textsc{clasp}, \cite{DBLP:journals/ai/GebserKS12}, \textsc{lp2sat} \cite{DBLP:journals/ki/Janhunen18a}, and \textsc{wasp} \cite{DBLP:conf/lpnmr/AlvianoDLR15}. 

\begin{table}[t!]
{
\renewcommand{\arraystretch}{1.1}
    \begin{tabular}{lcccc}
\hline
        Benchmark &  Choice Rules & Recursive Atoms & Count & Sum \\
\hline
        Graceful Graphs & \checkmark & & \checkmark &  \\
        Graph Colouring & & & & \\
        Hamiltonian &  \checkmark & \checkmark & \checkmark & \\
        Incremental Scheduling & \checkmark & & \checkmark & \checkmark \\
        Sokoban & \checkmark & & \checkmark & \\
\hline
    \end{tabular}
    \caption{\label{tab:featuresasp} List of ASP constructs available for each considered benchmark.}
}
\end{table}

We chose benchmark sets used in ASP competitions for which either a sufficiently large number of instances or a generator is available.
Selected benchmark sets include: Graceful Graphs, Graph Colouring, Hamiltonian, Incremental Scheduling, and Sokoban.
Concerning Incremental Scheduling and Sokoban, we use all the instances submitted to the 2015 ASP Competition~\cite{DBLP:journals/jair/GebserMR17}, whereas for Graceful Graphs, Graph Colouring and Hamiltonian instances were randomly generated.
Benchmarks were randomly divided into training and testing instances, aiming at having between 50 and 100 instances for testing purposes, and a 200-500 instances for training. In this setting, we considered a cutoff of 10 CPU-time minutes, and a memory limit of 8 GB of RAM.
Table~\ref{tab:featuresasp} summarises all the ASP constructs that are available for each tested benchmark.
Moreover, concerning \textsc{lp2sat}, there are several levels of configurations. In particular, it is possible \textit{(i)} to configure the ASP input and then to configure the SAT formula, \textit{(ii)} to configure only the SAT formula, \textit{(iii)} to configure only the ASP input. We have conducted a preliminary experiment analysis and in the following we report only the results of \textit{(iii)} since they are the ones that obtained the best performance.
Note that no domain contains (stratified) disjunction, so in this restricted setting ASP is as expressive as SAT (i.e., point (d) in the introduction is not leveraged). We are not aware of any publicly-available benchmark containing disjunctive rules with a huge number of instances and/or generators, therefore we could not extend our analysis to this kind of programs.


\begin{table}[t]
{
\renewcommand{\arraystretch}{1.1}
    \begin{tabular}{lrrrrrrr}
    & & \multicolumn{2}{c}{\textsc{clasp}} & \multicolumn{2}{c}{\textsc{lp2sat}} & \multicolumn{2}{c}{\textsc{wasp}} \\
\cline{3-4}\cline{5-6}\cline{7-8}
Problem & \# & Def. & Conf. & Def. & Conf. & Def. & Conf.\\
\cline{1-8}
Graph Colouring & 50 & 48  &  48 & 49  &  49 & 43  &  \textbf{44} \\
Graceful Graphs & 50 &  24  &  24 & 24   &  24 & 20  &  \textbf{22}\\
Hamiltonian & 50 & 50  &  50& 0  &  0 & 50  &  50 \\
Incremental Scheduling & 50 & 38  &  \textbf{39} & 25  &  25 & 36  &  \textbf{37} \\
Sokoban & 100 & 44  &   44 &  40  &   \textbf{44} & 40  &   \textbf{42}\\
\cline{1-8}
Total & 300 & 204  &  \textbf{205}  & 138  &  \textbf{142}  & 189  &  \textbf{195}  \\
\cline{1-8}
\end{tabular}
}
\caption{\label{tab:expASP1} Results of the selected solvers on the considered benchmark set. For each solver and benchmark, we show the number of solved instances when running on the default and on the configured ASP programs. Bold indicates the best result.} 
\end{table}

Table \ref{tab:expASP1} summarises the results of the selected solvers on the considered benchmark sets. Results are presented in terms of number of solved testing instances, achieved by solvers run using either the default or the configured ASP program. It is interesting to notice that, also for ASP, solvers are differently affected by the use of the configured knowledge. On the one hand, \textsc{clasp} does not benefit by the configuration in terms of coverage. On the other hand, the configuration has a widespread beneficial impact on the performance of \textsc{wasp}. \textsc{lp2sat} sits between the two sides of the spectrum: the configuration provided a significant improvement to the coverage performance on a single domain, Sokoban.


\begin{table}[t]
{
\renewcommand{\arraystretch}{1.1}
    \begin{tabular}{lrrrrrr}
    & \multicolumn{2}{c}{\textsc{clasp}} & \multicolumn{2}{c}{\textsc{lp2sat}} & \multicolumn{2}{c}{\textsc{wasp}} \\
\cline{2-3}\cline{4-5}\cline{6-7}
Problem & Def. & Conf. & Def. & Conf. & Def. & Conf.\\
\cline{1-7}
Graph Colouring & 47.3  &  \textbf{47.9} & 49.0  &  49.0 & 41.6  &  \textbf{44.0} \\
Graceful Graphs & 18.8  &  \textbf{22.3} & 24.0  &  24.0 & 18.6  &  \textbf{20.0}\\
Hamiltonian & 44.8  &  \textbf{46.0}& 0.0  &  0.0 & 50.0  &  50.0\\
Incremental Scheduling & 37.7  &  \textbf{38.9} & 24.6  &  \textbf{24.9} & 35.4  &  \textbf{35.6} \\
Sokoban & 44.0  &   44.0 &  37.6  &   \textbf{39.2} & 38.9  &   \textbf{40.5}\\
\cline{1-7}
Total & 192.6  &  \textbf{199.1}  & 135.2  &  \textbf{137.1}  & 184.5  &  \textbf{190.1}  \\
\cline{1-7}
\end{tabular}
\caption{\label{tab:expASP2} Results of the selected solvers on the considered benchmark sets. For each solver and benchmark, we show the IPC score achieved when running on the default and on the configured ASP programs. Bold indicates the best result.}
}
\end{table}
\begin{table}[t]
{
\renewcommand{\arraystretch}{1.1}
    \begin{tabular}{lrrrrrr}
    & \multicolumn{2}{c}{\textsc{clasp}} & \multicolumn{2}{c}{\textsc{lp2sat}} & \multicolumn{2}{c}{\textsc{wasp}} \\
\cline{2-3}\cline{4-5}\cline{6-7}
Problem & Def. & Conf. & Def. & Conf. & Def. & Conf.\\
\cline{1-7}
Graph Colouring & 262.3	& \textbf{261.7} 	& 4.6 	& 4.6 	& 850.8 	& \textbf{734.8} \\
Graceful Graphs & 3138.5 	& \textbf{3136.4}	& 3126.2 	& 3126.2 	& 3618.7 	& \textbf{3487.6}\\
Hamiltonian & 32.3 	& \textbf{26.3}	& 6000 	& 6000 	& 63.9 	& 63.9\\
Incremental Scheduling & 1466.1 	& \textbf{1352.7}	& 3020.6 	& \textbf{3015.3} 	& 2089.6 	& \textbf{1981.1} \\
Sokoban & 3401.2 	&  3401.2 	& 3621.0 	& \textbf{3410.7} & 3620.5 	& \textbf{3502.5}\\
\cline{1-7}
\end{tabular}
\caption{\label{tab:expASP3} Results of the selected solvers on the considered benchmark sets. For each solver and benchmark, we show the PAR10 score achieved when running on the default and on the configured ASP programs. Bold indicates the best result.}
}
\end{table}

In Tables \ref{tab:expASP2} and \ref{tab:expASP3} the performance of a solver run on the default and configured programs are compared in terms of IPC score and PAR10, respectively. Here it is possible to observe that the configuration has a beneficial impact on the considered solvers in most of the benchmark domains. When considering the performance of \textsc{clasp} in Hamiltonian, for instance, we observed an average runtime drop from 32 to 26 CPU-time seconds. 
Similar improvements have been observed also for the other solvers.  The presented results indicate that the configuration of ASP programs can improve the runtime performance of ASP solvers. 


Finally, we also test if there is a single general configuration that improves the performance of a solver on any ASP program, despite of the benchmark and underlying structure. As in the SAT counterpart, we trained each of the considered solvers on a training set composed by an equal proportion of instances from each of the benchmark sets. It should come as no surprise that the results indicate that this sort of configuration has no significant impact on ASP solvers' performance. For no one of the considered solvers it has been possible to identify a configuration able to improve the average performance. 

Also for ASP solvers, we used the fANOVA tool to identify the most relevant aspects of the configuration process. 
The lack of a general configuration that allows to improve average performance of the solvers, suggests that the importance of a configuration parameter depends on the benchmark domain. In other words, the same element can be more important in a domain, but almost irrelevant in another, according the the structure of the instances to be solved. 
Looking at the configured ASP program identified for \textsc{lp2sat} in the Sokoban domain, it appears that features related to occurrences of literals in heads (k1), and to occurrences of literals in bodies are among the most relevant criterion. In particular, the $5$ most important parameters identified by the fANOVA tool are: (k1), (k23), (k7), (k15), and (k13). 
By analysing the configurations obtained for \textsc{wasp} it is possible to derive that, for this solver, most of the important criteria are different than those of \textsc{lp2sat}. Considering a domain where the configuration allowed \textsc{wasp} to obtain a significant improvement, Graph Colouring, the $5$ most important parameters identified by the fANOVA tool are: (k16), (k1), (k13), (k14), and (k9). Only (k1), that focuses on the importance of the occurrences of a literal in heads, is shared between the configuration of \textsc{lp2sat} and \textsc{wasp}. This suggests that different solvers are more sensitive to different aspects of the ASP program, and therefore require different configurations.
%
%
To shed some light into the relevant parameters for the same solver across different domains, we analysed the configuration of \textsc{wasp} on the Graceful Graphs domain. In this case, the $5$ most important parameters identified by the fANOVA tool  are: (k8), (k2), (k9), (k18), (k7). In this case, it is easy to notice that (k1) is not deemed to be relevant for improving the performance of the solver, and only (k9) is relevant for \textsc{wasp} on both Graceful Graphs and Graph Colouring. It is worth reminding that the fANOVA analysis does not provide information on the actual value of the parameters, but only on the impact that a parameter can have on the performance of a solver. On this regards, we observed that also in cases where the same parameter is identified as very important by the fANOVA analysis, its best selected value by SMAC can be different in different domains. 

Note that it is difficult to find a high level explanation on why the solvers benefit of this different order, since they present a complex structure, where each (small) change in the order of input might have an impact on several different components.
However, in our view, this represents one of the strengths of our approach, since tools for automatic configuration might understand some hidden properties of the instances that are not immediately visible even to developers.

\paragraph{Discussion.}
ASP solvers are more complex than SAT solvers, since they have to deal with many additional aspects that are not present in SAT, such as the minimality property of answer sets, and additional constructs, as for example the aggregates.
Albeit this additional complexity, we observed that reordering is beneficial for both \textsc{lp2sat} and \textsc{wasp}, while the impact on the performance of \textsc{clasp} is less noticeable.
We conducted an additional analysis on the implementation of the solvers and we identified several reasons that can explain the results:
\begin{itemize}
    \item The considered solvers do not work directly on the input program, since they use a technique, called Clark's Completion, that basically produces a propositional SAT formula which is then used to compute the answer sets of the original program. In this context, the impact of the reorder might be mitigated by the additional transformation made by the solvers. 
    \item Data structures employed by the ASP solvers might deactivate some of the parameters used for the configuration. As an example, \textsc{wasp} uses a dedicated data structure for storing binary clauses which are then checked before other clauses, independently from their order in the input program. \textsc{clasp} extends this special treatment also to ternary clauses. Concerning SAT solvers, as far as we know, only \textsc{glucose} has a similar data structure, whereas \textsc{lingeling} and \textsc{cadical} have an efficient memory management of binary clauses but they do not impose an order among the clauses.
    \item Our tool considers the solvers as black boxes and uses their default configurations. Most of them employ parameters that were tuned on instances without specific reordering. Different results might be obtained by employing other configurations or by tuning the heuristics with instances processed by our reordering tool. However, from our perspective, the configuration tool can be already incorporated into grounders/preprocessors as an additional feature, to have a configured instance given to the solvers.
    \item Differently from other ASP and SAT solvers, the default configuration of \textsc{clasp} uses the Maximum Occurrence of clauses of Minimum size (MOMS) heuristic to initialize its heuristic parameters. This led to a more uniform behaviour when it is executed on the same instance with different orders. We observed that the behaviour of \textsc{wasp} is much more dependent on the order of the instances as confirmed in our experiment and as also shown in Section~\ref{sec:syntheticexperiment}.
\end{itemize}

\subsection{Synthetic Experiment}~\label{sec:syntheticexperiment}
In this section, we report the results of an experimental analysis conducted on a synthetic benchmark. Goal of the experiment is to give an explanation of the different performance between CLASP and WASP, and investigate the different qualitative results achieved in SAT and ASP. As a side effect, we show that it is possible to improve the performance of the ASP solver \textsc{wasp} using a proper ordering of the input program.\\

In particular, we focused on the following (synthetic) problem:
$$
\begin{array}{llr}
r_1: & \{in(i,j) \mid j \in \{1,\ldots,h\}\} \leftarrow & \forall i \in \{1, \ldots,p\}\\
r_2: & \leftarrow \#\mathit{count}\{in(i,j) \mid j \in \{1,\ldots,h\}\} \neq 1 & \forall i \in \{1, \ldots,p\}\\
r_3: & \leftarrow in(i_1,j), in(i_2,j), i_1 < i_2 & \forall i_1,i_2 \in \{1, \ldots, p\}, j \in \{1,\ldots, h\}\\
r_4: & col(i,c_1) \vee col(i,c_2) \vee \ldots \vee col(i,c_k) \leftarrow & \forall i \in \{1,\ldots, n\}\\
r_5: & \leftarrow edge(i_1,i_2), col(i_1,c), col(i_2,c) & \forall i_1, i_2 \in \{1,\ldots,n\}, c \in \{c_1,c_2,\ldots,c_k\}\\
r_6: & edge(i,j) \leftarrow & \forall i,j \in \{1,\ldots,n\}, i \neq j \\
\end{array}
$$
where rules $r_1$, $r_2$ and $r_3$ encode the pigeonhole problem with $p$ pigeons and $h$ holes, rules $r_4$ and $r_5$ encode the $k$-graph colouring problem, and $r_6$ encodes a complete graph with $n$ nodes provided as input to the graph colouring problem.
It is possible to observe that $r_1$, $r_2$, and $r_3$ admit no stable model when $p > h$, whereas $r_4$, $r_5$, and $r_6$ admit no stable model when $n > k$.
Concerning the pigeonhole problem, it is well-known that the performance of CDCL and resolution-based solvers are poor when $p > h$ and $p$ is greater than a given threshold~\cite{DBLP:series/faia/2009-185,DBLP:journals/tcs/Haken85}.
For instance, \textsc{clasp} terminates after 1.51 seconds when $h=9$ and $p=10$, after 17.81 seconds when
$h=10$ and $p=11$, and it does not terminate within 5 minutes when $h=11$ and $p=12$.
Similarly, concerning the $k$-graph colouring problem, large values of $k$ (e.g. $k \geq 10$) with $n > k$ are associated with poor performance of the solver.
Such properties are important in our case, since we are now able to control the hardness of the instances by properly selecting values of $h$, $p$, $k$, and $n$.
In particular, if the two sub-problems are combined, i.e., when the solver is executed on rules from $r_1$ to $r_6$, then we are able to create hard and easy sub-programs.
For instance, if we consider the case with $h=11$, $p=12$, $k=5$, and $n=100$, then we have that the rules from $r_1$ to $r_6$ admit no stable model, which is hard to prove for rules from $r_1$ to $r_3$ and easy to prove for rules from $r_4$ to $r_6$.
In this case the performance of the solver depends on the sub-program considered at hand.
As noted in Section~\ref{sec:exp}, the heuristic of \textsc{clasp} is not dependent on the ordering of the input program and, in this case, it is able to automatically focus on the easy subprogram. On the other hand, we observed that the performance of \textsc{wasp} depends on the processed order of the variables.
This behaviour of \textsc{wasp} can be explained by looking at its branching heuristic, which first selects literals with the lowest ids and then it focuses on the sub-problem related to such literals.
Clearly, this might lead to poor performance on the programs described above if the hard sub-problem is considered first.

In the following we show that the performance of \textsc{wasp} can be improved by performing an additional step after that the program has been configured, i.e., ids of the literals can be sorted according to the value of $O_l$ in descending order.
In particular, we report the results of an experimental analysis conducted on instances of the rules $r_1$--$r_6$, where $h=10$, $k=5$, $p=[20,\ldots,40]$ and $n=[7,\ldots,29]$.
Overall, we considered 483 instances, where 433 were used for the configuration and 50 were used for the testing.
Results show that \textsc{wasp} without configuration solves 33 instances out of 50 with a PAR10 equal to 2108.63, whereas \textsc{wasp} after the configuration solves 39 instances out of 50 with a PAR10 equal to 1351.94.

It is important to emphasise that the results are obtained without changing the implementation of the solver. Such changes might be directly included in the grounders (e.g. as additional parameter of the system \textsc{dlv}~\cite{DBLP:conf/lpnmr/AlvianoCDFLPRVZ17}) or in specific tools dedicated to preprocess the input programs.







\section{Related Work}
\label{sec:rel}



In both SAT and ASP there have been numerous papers where machine-learning-based configuration techniques based on features computation have been employed. Traditionally, such approaches aimed at modifying the behaviour of the solvers by either configuring their exposed parameters, or by combining different solvers into portfolios. In this work we consider an orthogonal perspective, where we configure the way in which the input knowledge, i.e., formula or program, is presented to the solver. This is done with the idea that the way in which instances are formulated and ordered can carry some knowledge about the underlying structure of the problem to be solved. In the following, we present main related literature in SAT and ASP, in two different paragraphs.

\paragraph{SAT.} 
A significant amount of work in the area has focused on approaches for configuring the exposed parameters of SAT solvers in order to affect their behaviour \cite{DBLP:journals/jair/EggenspergerLH19}. Well-known examples include the use of {\sc ParamILS} for configuring {\sc SAPS} and {\sc SPEAR} \cite{DBLP:journals/jair/HutterHLS09}, and of {\sc ReACTR} for configuring \textsc{lingeling} \cite{DBLP:conf/ijcai/FitzgeraldMO15}. Some approaches also looked into the generation of instance-specific configurations of solvers \cite{kadioglu2010isac}. This line of work led to the development of dedicated tools, such as {\sc SpySMAC} \cite{DBLP:conf/sat/FalknerLH15} and {\sc CAVE} \cite{DBLP:conf/lion/BiedenkappMLH18}, and to the organisation of the dedicated Configurable SAT Solver Challenge \cite{DBLP:journals/ai/HutterLBBHL17}. It also lead to the design and development of solvers, such as {\sc SATenstein} \cite{DBLP:journals/ai/KhudaBukhshXHL16}, that are very modular and natively support the use of configuration to combine all the relevant modules. A large body of works also focused on techniques for automatically configuring portfolios of solvers, such as {\sc satzilla} \cite{DBLP:journals/jair/XuHHL08}, based on the use of empirical prediction models of the performance of the considered solvers on the instance to be solved \cite{DBLP:conf/icml/HutterHL14}. Tools for assessing the contribution of different solvers to a portfolio has been introduced \cite{DBLP:conf/sat/XuHHL12}. With regard to portfolio generation, the interested reader is referred to \cite{DBLP:series/lncs/0001KMMO16}. 
The use of configuration techniques to generate portfolios of solvers has also been extensively investigated: {\sc HYDRA} \cite{DBLP:conf/aaai/XuHL10} builds a set of solvers with complementary strengths by iteratively configuring new algorithms; {\sc AutoFolio} \cite{DBLP:journals/jair/LindauerHHS15} uses algorithm selection to optimise the performance of algorithm selection systems by determining the best selection approach and its hyperparameters; finally, in \cite{DBLP:journals/ai/LindauerHLS17} an approach based on algorithm configuration for the automated construction of parallel portfolios has been introduced. The approach we follow relates also to the problem of discovering a backdoor, i.e., an ordering that will allow the problem to be solved faster, see, e.g. \cite{KilbySTW05}. However, it has to be noted that this is not a characteristic of our approach to configuration, but common to many approaches mentioned above.


\paragraph{ASP.}
Inspired by the solver \textsc{satzilla} in the area of SAT,
the \textsc{claspfolio} system \cite{DBLP:conf/lpnmr/GebserKKSSZ11,DBLP:journals/tplp/HoosLS14}
uses support vector regression to learn scoring functions approximating
the performance of several {\sc clasp} variants in a training phase.
Given an instance, {\sc claspfolio} then extracts features and evaluates such functions
in order to pick the most promising {\sc clasp} variant for solving the instance.
This algorithm selection approach was particularly successful in the Third ASP
Competition \cite{DBLP:journals/tplp/CalimeriIR14}, held in 2011,
where {\sc claspfolio} won the first place in the NP category
and the second place overall (without participating in the BeyondNP category).
The {\sc measp} system \cite{DBLP:journals/tplp/MarateaPR14,DBLP:conf/lpnmr/MarateaPR15}
goes beyond the solver-specific setting of {\sc claspfolio} and
chooses among different grounders as well as solvers.
Grounder selection traces back to \cite{DBLP:conf/aiia/MarateaPR13},
and similar to the QBF solver {\sc aqme} \cite{DBLP:journals/constraints/PulinaT09},
{\sc measp} uses a classification method for performance prediction.
Notably, ``bad'' classifications can be treated by adding respective instances
to the training set of {\sc measp} \cite{DBLP:journals/logcom/MarateaPR15},
which enables an adjustment to new problems or instances thereof.
Some of the parameters reported in Section~\ref{sec:configasp} were also adopted by \textsc{claspfolio} and \textsc{measp}, since they were recognised to be important to discriminate the properties of the input program.
In the Seventh ASP Competition~\cite{DBLP:journals/jair/GebserMR17}, the winning system was \textsc{i-dlv+s}~\cite{DBLP:journals/tplp/CalimeriDFPZ20} that utilises \textsc{ i-dlv}~\cite{DBLP:journals/ia/CalimeriFPZ17} for grounding and automatically selects back-ends for solving through classification between {\sc {\small clasp}} and {\sc {\small wasp}}.
Going beyond the selection of a single solving strategy from a portfolio,
the {\sc aspeed} system \cite{DBLP:journals/tplp/HoosKLS15} indeed runs different
solvers, sequentially or in parallel,
as successfully performed by {\sc ppfolio}.
Given a benchmark set, a fixed time limit per instance, and performance
results for candidate solvers,
the idea of {\sc aspeed} is to assign time budgets to the solvers such that
a maximum number of instances can be completed within the allotted time.
In other words, the goal is to divide the total runtime per computing core
among solvers such that the number of instances on which
at least one solver successfully completes its run is maximised.
The portfolio then consists of all solvers assigned a non-zero time budget
along with a schedule of solvers to run on the same computing core.
Calculating such an optimal portfolio for a benchmark set
is an Optimisation problem addressed with ASP in {\sc aspeed}. In \cite{DBLP:journals/corr/abs-2009-10240}, an approach for encoding selection was presented as a strategy for improving the performance of answer set solvers. In particular, the idea is to create an automated process for generating alternative encodings. The presented tool, called Automated Aggregator, was able to handle non-ground count aggregates. Other automatic non-ground program rewriting techniques are presented in \cite{DBLP:conf/padl/HippenL19}.\\

Further, there has been a recent growing interest in techniques for knowledge model configuration in the areas of AI Planning and Abstract Argumentation. In AI Planning, it has been demonstrated that the configuration of either the domain model \cite{DBLP:conf/ijcai/VallatiHCM15,DBLP:journals/jar/VallatiCMH21} or the problem model \cite{DBLP:conf/aips/VallatiS18} can lead to significant performance improvement for domain-independent planning engines. On the argumentation side, it has been shown that even on syntactically simple models that represent directed graphs, the configuration process can lead to performance improvements \cite{DBLP:journals/ijar/CeruttiVG18}.

\section{Conclusions}
\label{sec:conc}


In this paper we proposed an approach for exploiting the fact that the order in which the main elements of CNF formulae and ASP programs, i.e., literals, clauses and rules, are listed carries some information about the structure of the problem to be solved, and therefore affect the performance of solvers. The proposed approach allows to perform the automated configuration of formulae and programs. In SAT, we considered as configurable the order in which clauses are listed and the order in which literals are listed in the clauses, while for ASP we considered as configurable similar entities, i.e., literals and rules, but taking into account that some rule's structure has to be maintained, and that other powerful constructs like aggregates come into play. In our experimental analysis we configured formulae and programs for improving number of solved instances and PAR10 performance of solvers. 
The performed analysis, aimed at investigating how the configuration affects the performance of state-of-the-art solvers: (i) demonstrates that the automated configuration has a significant impact on solvers' performance, more evident for SAT but also significant for ASP, despite the constraints on rule's structure; (ii) indicates that the configuration should be performed on specific set of benchmarks for a given solver; and (iii) highlights what are the main features and aspects of formulae and programs that have a potentially strong impact on the performance of solvers. Such features can be taken into account by knowledge engineers to encode formulae or programs in a way that supports solvers that will reason upon them. 

Our findings can have implications on both solving and encoding side. Given the positive results obtained, solver's developers should consider rearranging internally the structure of the formula/program in order to optimise their performance, and/or users could their-selves consider these hints while writing the encoding. However, given that such positive results are obtained per-domain, and varies with solvers, care should be taken in doing so.  


We see several avenues for future work. 
We plan to evaluate the impact of configuration on optimisation variants of SAT and ASP, i.e., weighted max SAT, or ASP including soft constraints, where the weight of the elements can provide another important information to the configuration process. We are also interested in evaluating if ordering clauses (and literals) that are learnt during the search process of a SAT solver can be beneficial for improving performance, given that all of the solvers employed are based on (variant of) the CDCL algorithm.
In this paper we focused on exact solvers based on CDCL, as future work it can be interesting to consider also SAT solvers based on local search. Concerning ASP solvers, it can be also interesting to check if reordering the non-ground program can have a positive impact on the performance. This would also open the combination of reordering and systems based on lazy-grounding~\cite{DBLP:journals/tplp/WeinzierlTF20}. 
Another interesting future work can be to investigate the joint tuning of ordering and solver parameters.
Moreover, in this paper we focused on parameters that are based on our knowledge of existing solvers, involving parameters, among others, that proved to be important to characterised input programs; of course, the inclusion of additional parameters is possible, e.g., taking into account theoretical properties of the input programs (see, e.g., \cite{DBLP:journals/tplp/FichteTW15,DBLP:journals/jancl/Janhunen06,DBLP:journals/tocl/LifschitzPV01,DBLP:journals/tocl/GebserS13}), or adding symmetric counterparts of current criteria, e.g., (c6) and (c7) of the input formula. 
Finally, we plan to incorporate the re-ordering into existing approaches for configuring portfolios of SAT solvers, such as SATenstein, which works in a similar way as {\sc aspeed}, but differently from {\sc aspeed} is not tailored on different configurations of the same solver, in order to further improve performance, and to investigate the concurrent configuration of formulae/programs and solvers. 

\paragraph{Competing interests declaration.} The authors declare none.

\bibliographystyle{acmtrans}
\bibliography{biblio}

\label{lastpage}
\end{document}